\documentclass[lettersize,journal]{IEEEtran}
\usepackage{amsmath,amsfonts,amssymb}
\usepackage{algorithmic}
\usepackage{algorithm}
\usepackage{array}
\usepackage[caption=false,font=normalsize,labelfont=sf,textfont=sf]{subfig}
\usepackage{textcomp}
\usepackage{stfloats}
\usepackage{url}
\usepackage{verbatim}
\usepackage{hyperref}
\usepackage{graphicx}
\usepackage{booktabs}
\usepackage{cite}
\usepackage{tikz}
\usetikzlibrary{matrix}
\usepackage{subcaption}
\usepackage{tabularx}
\usepackage{adjustbox}

\begin{document}

\title{Efficient and Generalizable Archetypal Analysis for Discrete Data}

\author{A. Emilie J. Wedenborg, Jesper Løve Hinrich and Morten Mørup
\thanks{A. E. J. Wedenborg, J. L. Hinrich, and M. M{\o}rup are with the Department of Applied Mathematics and Computer Science, Technical University of Denmark, 2800 Kgs. Lyngby, Denmark.} \thanks{Corresponding author: A. Emilie J. Wedenborg (e-mail: aejew@dtu.dk).}
\thanks{A preliminary version of part of this work was presented at the 2025 IEEE International Conference on Acoustics, Speech and Signal Processing (ICASSP) as "Archetypal Analysis for Binary Data" \cite{WedenborgBinary}.}%
\thanks{This work was supported by the Danish Data Science Academy, which is funded by the Novo Nordisk Foundation (NNF21SA0069429) and VILLUM FONDEN (40516).}
\thanks{This work has been submitted to the IEEE for possible publication. Copyright may be transferred without notice, after which this version may no longer be accessible.}}

\maketitle

\begin{abstract}
Archetypal Analysis (AA) represents observations as convex combinations of extremal data-driven profiles, yielding interpretable low-dimensional descriptions of complex datasets. Classical AA is typically formulated with a least-squares objective, which is poorly matched to binary, count, and categorical observations arising in domains such as genomics, microbiome analysis, recommender systems, and medical questionnaires. We introduce an efficient and generic likelihood-based framework for AA of discrete data that supports Bernoulli, Poisson, and multinomial observation models. The proposed optimization scheme uses local quadratic approximations of the associated AA-models negative log-likelihood, enabling efficient constrained updates through sequential minimal optimization (SMO) for observation-specific mixture weights and an active-set method for mixture weights used to construct each archetype from the data observations. To improve scalability, the active set is explicitly bounded while maintaining feasibility through a simplex-preserving proxy representation. We further propose a cross-validated predictive likelihood framework relying on quantifying how the trained AA representations generalize to held-out data as a principled criterion for selecting the number of archetypes, providing an alternative to reconstruction-error heuristics and stability-only diagnostics. Synthetic experiments establish the efficiency of our proposed procedure and show that the proposed predictive likelihood framework recovers the true model complexity. Applications to single-cell RNA sequencing, microbiome composition, and somatic mutation data further demonstrate that the learned archetypes capture interpretable domain-specific structures while providing competitive likelihood fits and stable solutions when compared to alternative inference procedures. Notably, the predictive likelihood framework here also points to suitable model complexities guiding the model selection. These results establish the proposed AA framework as an efficient approach to likelihood-based inference in AA for discrete data with the associated predictive likelihood based approach providing a complementary new tool for model order selection in AA.
\end{abstract}

\begin{IEEEkeywords}
Archetypal analysis, discrete data, likelihood-based optimization, sequential minimal optimization, active-set methods, predictive likelihood, model selection
\end{IEEEkeywords}

\section{Introduction}

Understanding complex datasets often requires identifying the most extreme and informative patterns that govern variability. Rather than partitioning data into clusters or projecting observations onto orthogonal latent components, Archetypal Analysis (AA), originally introduced by \cite{Cutler01111994}, provides a geometric framework in which each data point is expressed as a convex combination of extremal learned representatives. AA models observations as mixtures of archetypes that themselves lie within the convex hull of the dataset. This construction induces a low-dimensional polytope embedded in the feature space, where vertices correspond to optimally distinctive data-driven profiles.

This perspective distinguishes AA from clustering approaches, which assign observations to discrete groups with the extracted prototypes resembling average properties of the groups, and from classical matrix factorization techniques such as non-negative matrix factorization \cite{Lee2000AlgorithmsFactorization}, which do not enforce extremality of the latent factors but produces part based representations\cite{lee1999learning}, see also\cite{Mrup2012ArchetypalMining}. By constraining archetypes to be convex combinations of observed samples, AA yields interpretable representations that capture trade-offs between extreme yet realizable configurations. As a result, AA has found applications across neuroscience \cite{Hinrich2016ArchetypalData,Olsen2022CombiningAnalysis}, genomics \cite{Thgersen2013ArchetypalAirways}, computer vision \cite{Chen2014/CVPR.2014.192,Wynen2018UnsupervisedAnalysis}, and deep generative modeling \cite{Keller2019DeepAnalysis}, see  \cite{alcacer2025survey} for a recent survey.

\subsection{Optimization Challenges in Archetypal Analysis}

Despite its interpretability, AA presents substantial computational challenges. The estimation problem is inherently non-convex, as both the convex mixtures used to define the archetypes and the convex reconstruction weights must be learned simultaneously. Standard approaches therefore alternate between two convex subproblems formed by solving for each set of parameters at a time, typically formulated as quadratic programs that converge to a local optimum.

Numerous algorithmic refinements have been proposed to improve scalability and stability. The Principal Convex Hull Algorithm (PCHA) introduced by \cite{Mrup2012ArchetypalMining} provides efficient projected-gradient updates and remains a widely used baseline. Other work has explored robust loss formulations \cite{Chen2014/CVPR.2014.192}, coreset constructions for large-scale data \cite{Mair2019CoresetsAnalysis}, probabilistic relaxations \cite{Han2022ProbabilisticAnalysis}, and deep-learning-based extensions \cite{Keller2019DeepAnalysis}. While these advances significantly improve performance for continuous data under squared-error objectives, most of these existing optimization strategies remain tightly coupled to least-squares reconstruction.

\subsection{Archetypal Analysis for Discrete Data}

In many modern applications, including genomics, recommender systems, medical questionnaires, and single-cell sequencing, observations are binary, categorical, or counts. This makes a classical least-squares AA method suboptimal. 

Probabilistic extensions of AA (PAA) have attempted to address this limitation \cite{Seth2016ProbabilisticAnalysis}. For binary and count data, multiplicative update rules inspired by non-negative factorization methods have been proposed. However, these updates are often slow to converge and sensitive to initialization \cite{Gillis2008NonnegativeProblem,Nielsen2014Non-negativeBrain}. Alternative strategies rely on transforming discrete data into continuous representations \cite{Gimbernat-Mayol2021ArchetypalGenetics} or constraining archetypes to coincide with actual observations (archetypoids) \cite{CaberoFayos2020FindingQuestionnaires}. While useful, such approaches either distort the underlying data-generating process or restrict model flexibility.

Consequently, there remains a need for a computationally efficient and statistically principled optimization framework for likelihood-based AA applicable to discrete data.

\subsection{Proposed Framework}

We develop a unified and efficient likelihood-based inference framework for AA, named Distribution Agnostic Archetypal Analysis (DAAA). Our approach is based on three key principles:

\paragraph{Exploiting sparsity in archetype construction}
The convex combinations defining the archetypes are typically sparse at optimality. We exploit this property using an active-set strategy inspired by classical non-negativity constrained least squares methods \cite{Bro1997AAlgorithm}, enabling efficient updates under simplex constraints. To further promote computational efficiency, we explicitly control the size of the active set by imposing a bound on the active set size using a reparameterization in which the current archetypal estimate is used as an additional observation. When the active set exceeds this limit, it is thereby reset to a single element pointing to this pseudo-observation whereas the weights are reassigned in a manner that preserves feasibility and ensures convergence. This strategy guarantees that the active set procedure remains computationally lightweight while maintaining strict convergence.

\paragraph{Low-dimensional reconstruction via Sequential Minimal Optimization}
The observation-specific convex weights typically lie in a low-dimensional simplex defined by the number of archetypes used for the modeling. We formulate their update as a sequence of analytically solvable two-variable subproblems using Sequential Minimal Optimization (SMO) \cite{Platt1998SequentialMachines}, yielding closed-form updates for these two-variable subproblems with low computational overhead.

\paragraph{Second-order likelihood expansion}
Rather than deriving individual solvers for each observation model as in \cite{Seth2016ProbabilisticAnalysis}, we construct a quadratic approximation of the negative log-likelihood at each iteration. This transforms the general likelihood-based objective into a sequence of weighted quadratic subproblems compatible with our active-set and SMO updates. We focus on discrete observation models, for which the likelihood-specific gradient and second-order (curvature) terms admit simple expressions. However, this approach is generic and we note that the optimization strategy applies to any twice-differentiable separable loss for which a local quadratic approximation can be constructed.

\subsection{Model Selection via Predictive Likelihood}

Determining the number of archetypes $K$ remains a longstanding open problem in AA. Existing approaches typically rely on heuristic inspection of reconstruction error curves \cite{Mrup2012ArchetypalMining} or stability measures such as normalized mutual information \cite{Hinrich2016ArchetypalData}. Alternatives have been information criteria relying on training sample estimates \cite{prabhakaran2012automatic,suleman2017validation}, see also \cite{alcacer2025survey}.
We presently introduce an out-of-sample predictive likelihood framework for selecting $K$ based on evaluating the held-out (predictive) log-likelihood (LL) of the trained archetypal model. We thereby transform archetype selection into a principled problem optimizing for generalization. This thereby bridges the AA model selection problem with established statistical learning methodology and provides a direct model generalization criteria for determining the complexity of the archetypal analysis induced polytope.

\subsection{Contributions}

In summary, our  main contributions are:

\begin{itemize}
    \item A likelihood-based AA inference framework with an optimization strategy that extends to general twice-differentiable separable losses through local quadratic approximations highlighted in the context of AA for discrete data.
    \item Closed-form efficient inference for the two convex subproblems, respectively exploring a constrained active-set procedure to bound the update complexity for the typically sparse mixture weights defining the archetypes, and a SMO procedure for the typical low-dimensional simplex structure used to reconstruct each observation.
    \item A predictive likelihood criterion for principled selection of the number of archetypes considering out-of-sample model generalization.
    \item Extensive empirical validation on synthetic and real-world datasets demonstrating improved convergence speed and statistical performance compared to multiplicative-update methods and projected gradient extending presently also the PCHA procedure to AA on discrete data.
\end{itemize}

\section{Methods}

\subsection{Likelihood-Optimized Archetypal Analysis}

We introduce a general likelihood-based formulation of AA \cite{Cutler01111994}. Let $\mathbf{X}\in \mathbb{R}^{M\times N}$ denote a data matrix with $M$ features and $N$ observations. The goal is to represent each observation as a convex combination of $K$ archetypes, while constraining each archetype to lie in the convex hull of the observed data.

Let $\mathbf{C}\in \mathbb{R}^{N\times K}$ and $\mathbf{S}\in \mathbb{R}^{K\times N}$ be column-stochastic matrices. The archetype matrix is $\mathbf{Z} = \mathbf{XC}$, and the reconstructed parameter matrix is $\mathbf{R} = \mathbf{ZS} = \mathbf{XCS}$. The columns of $\mathbf{C}$ define the archetypes as convex combinations of observations, whereas the columns of $\mathbf{S}$ define each observation as a convex combination of archetypes.

We estimate $\mathbf{C}$ and $\mathbf{S}$ by minimizing a negative log-likelihood subject to the constraints that the columns of $\mathbf{C}$ and $\mathbf{S}$ reside on the standard simplex:

\begin{equation} \label{eq:AAClassic}
\begin{aligned}
\min_{\mathbf{C}, \mathbf{S}} \quad & L(\mathbf{X}, \mathbf{R}) \\
\text{s.t.} \quad 
& \mathbf{R} = \mathbf{X}\mathbf{C}\mathbf{S}, \\
& c_{j,k} \ge 0, \quad s_{k,j} \ge 0, \\
& \sum_{j=1}^{N} c_{j,k} = 1, \quad 
  \sum_{k=1}^{K} s_{k,j} = 1 ,
\end{aligned}
\end{equation}

The choice of loss function $L(\mathbf{X}, \mathbf{R})$ corresponds to choice of likelihood used to statistically model the data and allows AA to be adapted to different data types. In particular, likelihood-based formulations provide a principled way to tailor AA to specific observation models, as discussed in \cite{Seth2016ProbabilisticAnalysis}. 

The classical formulation corresponds to a least-squares objective,
\begin{equation}
\label{eq:LSloss}
L_{\text{LS}}(\mathbf{X}, \mathbf{R}) 
= \sum_{i=1}^{M} \sum_{j=1}^{N} 
\left(x_{i,j} - r_{i,j}\right)^2,
\end{equation}
which is equivalent to maximum likelihood estimation assuming independent Gaussian noise.

More generally, by defining $L$ as the negative log-likelihood (NLL), AA can be interpreted as a constrained likelihood maximization problem. This likelihood-based perspective forms the foundation of the optimization framework developed in the following sections. Notably, the problem in eq. \eqref{eq:AAClassic} is jointly non-convex in $(\mathbf{C},\mathbf{S})$ but convex when optimizing $\mathbf{C}$ keeping $\mathbf{S})$ fixed and vice versa. Therefore, we adopt the standard alternating optimization strategy for AA in which $\mathbf{C}$ and $\mathbf{S}$ are updated iteratively while holding the other fixed until a local maximum is reached. 

\subsection{Quadratic Approximation of the Likelihood}
To obtain efficient updates for likelihood-based AA,
we approximate the NLL locally by a second-order Taylor expansion. This converts each alternating subproblem into a convex quadratic program under simplex constraints. The resulting quadratic form can then  be efficiently optimized using the SMO procedure for the columns of $\mathbf{S}$ and a constrained active-set procedure for the columns of $\mathbf{C}$.

For fixed $\mathbf{C}$, the reconstruction can be written as
\begin{equation}
    \mathbf{R} = \mathbf{ZS},
\qquad \mathbf{Z} = \mathbf{XC},
\end{equation}

and assuming the loss decomposes over observations,
\begin{equation}
    L(\mathbf{X},\mathbf{R}) = \sum_{j=1}^N \ell(\mathbf{x}_j, \mathbf{r}_j),
\end{equation}
then, when  $\mathbf{C}$ is fixed, the optimization over $\mathbf{S}$ separates into independent column-wise problems,
$f(\mathbf{s}_j)$. A second-order Taylor expansion around the current
iterate gives
\begin{equation}
\label{eq:Fs}
f(\mathbf{s}_j) 
\approx \text{const.} 
- \mathbf{d}_j^\top \mathbf{s}_j 
+ \frac{1}{2} \mathbf{s}_j^\top \mathbf{H}^{(j)} \mathbf{s}_j,
\end{equation}

Similarly, we can formulate a similar problem by fixing $\mathbf{S}$ and optimizing the columns in $\mathbf{C}$

\begin{equation} \label{eq:Fc}
f(\mathbf{c}_k) \approx \text{const.} -\mathbf{d}_k^\top\mathbf{c}_k+\frac{1}{2}\mathbf{c}_k^\top\mathbf{H}^{(k)}\mathbf{c}_k,
\end{equation}

\subsection{Updating S through sequential minimal optimization}

Originally proposed by \cite{Platt1998SequentialMachines} and adapted for AA in our related work \cite{WedenborgBinary}, SMO functions by decomposing large-scale Quadratic Programming (QP) problems into the smallest possible sub-problems.  This approach is particularly effective for updating the matrix $\mathbf{S}$, which is typically a low-dimensional dense matrix that represents the data as convex combinations of archetypes.  

The SMO framework optimizes archetypes in pairs, updating them iteratively to ensure global convergence. For any given pair of archetypes $k^{\prime}$ and $k^{\prime \prime}$, we define a joint weight $t_j$ as $t_j=\sum_{d \in \{k^\prime, k^{\prime\prime}\}} s_{d,j}$. To solve the sub-problem, we introduce a weight parameter $\alpha_j  \in [0,1]$. This allows us to re-parameterize the individual weights as $s_{k^{\prime},j}=t_j \alpha_j$ and $s_{k^{\prime\prime},j}=t_j (1-\alpha_j)$.
By substituting these into the loss function, the optimization reduces to finding the optimal $\alpha_j$ for each pair, significantly reducing the computational overhead of the original QP problem. Consequently, by substituting these parameters, the loss function $F(\mathbf{s}_j)$ can be approximated as a quadratic function of $\alpha_j$:

\begin{equation} \label{eq:Fs_matrix}
    \begin{aligned}
        F(\mathbf{s}_j) \approx const - \begin{bmatrix}
            d_{k^{\prime},j} \\
            d_{k^{\prime\prime},j}
        \end{bmatrix}^\top \begin{bmatrix}
             t_j \alpha_j \\
             t_j(1-\alpha_j)
        \end{bmatrix}+\\\frac{1}{2}
        \begin{bmatrix}
             t_j\alpha_j \\
             t_j(1-\alpha_j)
        \end{bmatrix}^\top \begin{bmatrix}
            h^{(j)}_{k^{\prime},k^{\prime}} & h^{(j)}_{k^{\prime},k^{\prime\prime}}\\            h^{(j)}_{k^{\prime\prime},k^{\prime}} & h^{(j)}_{k^{\prime\prime},k^{\prime\prime}}
        \end{bmatrix}
        \begin{bmatrix}
            t_j\alpha_j \\
            t_j(1-\alpha_j)
        \end{bmatrix}.
    \end{aligned}
\end{equation}

Since the loss function is quadratic with respect to $\alpha_j$, it produces the following closed-form update:
\begin{equation} \label{eq:alpha}
    \alpha^{\star}_j = -\frac{t_j h^{(j)}_{k^{\prime},2}+ \sum t_jh^{(j)}_{k^{\prime\prime},k^{\prime}}-2 t_j h^{(j)}_{k^{\prime\prime},k^{\prime\prime}}-2d_{k^{\prime},j}+2 d_{k^{\prime\prime},j}}{2t_j (h^{(j)}_{k^{\prime},k^{\prime}}-h^{(j)}_{k^{\prime},k^{\prime\prime}}-h^{(j)}_{k^{\prime\prime},k^{\prime}}+h^{(j)}_{k^{\prime\prime},k^{\prime\prime}})}.
\end{equation}

To maintain the validity of the convex combination, $\alpha^{\star}_j$ is restricted to the interval $[0,1]$. Given the convexity of the loss function, we apply a simple clipping: if $\alpha^{\star}_j> 1$, it is set to 1, and if $\alpha^{\star}_j<0$, it is set to 0.

\subsection{C-update by constrained active-set optimization}

For fixed $\mathbf{S}$, we update the archetype construction matrix $\mathbf{C}$. Since each archetype is defined as a convex combination of observations, the columns of $\mathbf{C}$ are constrained to lie on the simplex. Notably, only a relative few number of observations are typically here used to define each archetype. We exploit this expected sparsity of the convex combinations using an active-set strategy adapted from fast non-negative
least squares \cite{Bro1997AAlgorithm}.

For a single column $\mathbf{c}_k$, a second-order Taylor expansion of the likelihood gives the local quadratic approximation that can be seen in \autoref{eq:Fc}. To ensure that the Hessian remains full rank and numerically stable, a small regularization term $\frac{1}{2}\epsilon\sum_jc_{jk}^2$ is added.
\begin{equation} \label{eq:Fc_w_noise}
f(\mathbf{c}_k) \approx \text{const.} -\mathbf{d}_k^\top\mathbf{c}_k+\frac{1}{2}\mathbf{c}_k^\top(\mathbf{H}^{(k)}+\epsilon\mathbf{I})\mathbf{c}_k,
\end{equation}
where \(\mathbf{d}_k\) and \(\mathbf{H}^{(k)}\) denote likelihood-specific first- and second-order terms with respect to $\mathbf{c}_k$.

Rather than imposing the simplex constraint through a quadratic penalty \cite{Cutler01111994}, we enforce it directly as an equality constraint within the active set \cite{Bell2007ModelingSystems}. Let $A$ denote the current active set. The restricted update is obtained by solving the KKT system
\begin{equation}
\begin{bmatrix}
\mathbf{H}^{(k)}_{A,A} + \epsilon \mathbf{I} & \mathbf{1} \\
\mathbf{1}^{\top} & 0
\end{bmatrix}
\begin{bmatrix}
\mathbf{z} \\
\mu
\end{bmatrix}
=
\begin{bmatrix}
\mathbf{d}_{A} \\
1
\end{bmatrix},
\label{eq:C_update_KKT}
\end{equation}
where $\mathbf{z}$ contains the active coefficients, $\mu$ is the Lagrange multiplier for the sum-to-one constraint, and $\epsilon$ is a small ridge term used for numerical stability. Coefficients outside the active set are set to zero, and the active set is updated according to the standard non-negativity conditions.

To control the computational cost in high-dimensional settings, we restrict the active set size to at most $N_{\max}$. Suppose that, during the update of column $K$, the current coefficient vector is $\mathbf{c}_k\in\Delta_n$, where
\begin{equation}
\Delta_n=\{\mathbf{c}\in\mathbb{R}^n:\mathbf{c}\geq 0,\ \mathbf{1}^\top
\mathbf{c}=1\}.
\end{equation}
When the active set becomes too large, we replace the current active combination with a single proxy variable. More precisely, we define $\boldsymbol{\theta} = \mathbf{c}_k$. $\boldsymbol{\theta} \in \Delta_n$ can therefore be seen as a convex combination of the original observations, In data space, this corresponds to introducing the surrogate observation $\tilde{\mathbf{x}}_{\boldsymbol{\theta}} =\sum_{i=1}^{n}\theta_i\mathbf{x}_i, $ i.e. the observation obtained from the current active convex combination. The active set is then reset and initialized with this surrogate observation with
coefficient $c_{\mathrm{proxy}}=1$. Subsequent updates optimize using both the original observations and the proxy. If $A$ denotes the set of currently active original observations after the reset, the coefficient vector is represented as 

\begin{equation}
\mathbf{c}_k = \sum_{i\in A} c_i\mathbf{e}_i +
c_{\mathrm{proxy}}\boldsymbol{\theta}.
\label{eq:proxy_representation}
\end{equation}
Since both $\mathbf{e}_i$ and $\boldsymbol{\theta}$ lie in the simplex, the
simplex constraint is preserved by imposing
\begin{equation}
\sum_{i\in A}c_i+c_{\mathrm{proxy}}=1,
\quad
c_i\geq 0,\quad c_{\mathrm{proxy}}\geq 0.
\end{equation}
The KKT system is updated accordingly. Let 
\begin{equation}
    \mathbf{B}_A=
\begin{bmatrix}
\mathbf{e}_{i_1} & \cdots & \mathbf{e}_{i_{|A|}} &
\boldsymbol{\theta}
\end{bmatrix},
\end{equation}
where $A=\{i_1,\ldots,i_{|A|}\}$, and let
\begin{equation}
\boldsymbol{\alpha} =
(c_{i_1},\ldots,c_{i_{|A|}},c_{\mathrm{proxy}})^\top .
\end{equation}
Then $\mathbf{c}_k = \mathbf{B}_A\boldsymbol{\alpha}$ and the restricted quadratic program becomes 
\begin{equation}
\begin{aligned}
\min_{\boldsymbol{\alpha}\geq 0}\quad
&-\left(\mathbf{B}_A^\top\mathbf{d}_k\right)^\top\boldsymbol{\alpha}
+\frac{1}{2}\boldsymbol{\alpha}^\top \mathbf{B}_A^\top
\left(\mathbf{H}^{(k)}+\epsilon\mathbf{I}\right)\mathbf{B}_A\boldsymbol{\alpha} \\
\text{s.t.}\quad
&\mathbf{1}^\top\boldsymbol{\alpha}=1.
\end{aligned}
\end{equation}

The corresponding KKT system is 
\begin{equation}
    \begin{bmatrix}
\mathbf{B}_A^\top
\left(\mathbf{H}^{(k)}+\epsilon\mathbf{I}\right)
\mathbf{B}_A
&
\mathbf{1}
\\
\mathbf{1}^\top & 0
\end{bmatrix}
\begin{bmatrix}
\boldsymbol{\alpha}\\
\mu
\end{bmatrix}
=
\begin{bmatrix}
\mathbf{B}_A^\top\mathbf{d}_k\\
1
\end{bmatrix}.
\label{eq:C_update_KKT_proxy}
\end{equation}
Equivalently, the proxy contributes one additional row and column to the
restricted Hessian, with entries
\begin{equation}
\mathbf{H}^{(k)}_{A,\theta} = \mathbf{H}^{(k)}_{A,:}\boldsymbol{\theta}, \quad H^{(k)}_{\theta,\theta}
= \boldsymbol{\theta}^\top \mathbf{H}^{(k)} \boldsymbol{\theta},
\quad
d_{\theta} = \boldsymbol{\theta}^\top\mathbf{d}_k,
\end{equation}
with the same augmentation applied to the ridge term. Consequently, the proxy is not an
unconstrained approximation, it is an additional simplex-preserving surrogate observation whose effect on the quadratic subproblem is computed from the original likelihood expansion.

\subsection{Likelihood specific derivatives}
For all models, updates follow a generalized structure using the partial derivatives of the loss with respect to the model parameters $\mathbf{S}$ and $\mathbf{C}$ via the chain rule (tables \ref{tab:likelihood_factors} and \ref{tab:generic_update_components}).

When updating the matrix $\mathbf{S}$ for a fixed $\mathbf{C}$, we define the Jacobian term $\alpha_{i,k} = \frac{\partial r_{i,j}}{\partial s_{k,j}} = \sum_m p_{i,m} c_{m,k}$, with 
\begin{equation}
p_{i,j} =
\begin{cases}
x_{i,j}, & \text{(Frobenius)} \\
x_{i,j} + \varepsilon, & \text{(Poisson)} \\
x_{i,j}(1 - 2\varepsilon) + \varepsilon, & \text{(Bernoulli)}
\end{cases}
\end{equation}
or
\begin{equation}
p_{i,j,p} = x_{i,j,p}(1-\varepsilon) + \dfrac{\varepsilon}{L-1}(1 - x_{i,j,p}), \quad \text{(Multinomial)}
\end{equation}

The full update direction $d_{k,j}$ incorporates the partial derivative and an adjustment term:
\begin{equation}
    d_{k,j} = \frac{\partial L}{\partial s_{k,j}} - \sum_{k',k''} h^{(j)}_{k',k''} s_{k,j}
\end{equation}
where the partial derivative $\frac{\partial L}{\partial s_{k,j}}$ and the Hessian-diagonal approximation $h^{(j)}_{k',k''} = \frac{\partial^2 L}{\partial s_{k',j} \partial s_{k'',j}}$ are defined respectively for the Gaussian, Bernoulli, Poisson, and Multinomial likelihoods in table \ref{tab:likelihood_factors} and \ref{tab:generic_update_components}.

For the update of matrix $\mathbf{C}$ with fixed $\mathbf{S}$, we define the Jacobian term $\beta_{i,j,j',k} = \frac{\partial r_{i,j'}}{\partial c_{j,k}} = p_{i,j} s_{k,j'}$. 

The update direction $d_{j,k}$ is defined as:
\begin{equation}
    d_{j,k} = \frac{\partial L}{\partial c_{j,k}} + \sum_{j',j''} h^{(k)}_{j',j''} c_{j,k}
\end{equation}
with the respective partial derivatives and Hessian approximations $h^{(k)}_{j',j''} = \frac{\partial^2 L}{\partial c_{j',k} \partial c_{j'',k}}$ provided in Table \ref{tab:likelihood_factors} and \ref{tab:generic_update_components}.

\begin{table}[ht]
\centering
\scriptsize
\caption{Likelihood-specific gradient and Hessian factors.}
\label{tab:likelihood_factors}
\setlength{\tabcolsep}{4pt}
\renewcommand{\arraystretch}{1.25}

\begin{tabular}{@{}lll@{}}
\toprule
\textbf{Likelihood} 
& \textbf{Gradient factor } $g_{i,j}$ 
& \textbf{Hessian factor } $w_{i,j}$ \\
\midrule

Gaussian
& $-2(x_{i,j}-r_{i,j})$
& $2$ \\

Bernoulli
& $-\left(\dfrac{x_{i,j}}{r_{i,j}}-\dfrac{1-x_{i,j}}{1-r_{i,j}}\right)$
& $\dfrac{x_{i,j}}{r_{i,j}^2}+\dfrac{1-x_{i,j}}{(1-r_{i,j})^2}$ \\

Poisson
& $-\left(\dfrac{x_{i,j}}{r_{i,j}}-1\right)$
& $\dfrac{x_{i,j}}{r_{i,j}^2}$ \\

Multinomial
& $-\dfrac{x_{i,j,p}}{r_{i,j,p}}$
& $\dfrac{x_{i,j,p}}{r_{i,j,p}^2}$ \\

\bottomrule
\end{tabular}

\vspace{1ex}

\footnotesize
For the Multinomial likelihood, replace the feature index $j$ by the feature-category pair $(j,p)$.
\end{table}

\begin{table}[ht]
\centering
\scriptsize
\caption{Generic S- and C-update components using the likelihood-specific factors from Table~\ref{tab:likelihood_factors}.}
\label{tab:generic_update_components}
\setlength{\tabcolsep}{4pt}
\renewcommand{\arraystretch}{1.25}

\begin{tabular}{@{}lll@{}}
\toprule
\textbf{Update} 
& \textbf{Partial derivative} 
& \textbf{Hessian term} \\
\midrule

$\mathbf{S}$
& $\displaystyle \frac{\partial L}{\partial s_{k,j}}
= \sum_i g_{i,j}\alpha_{i,k}$
& $\displaystyle h^{(j)}_{k',k''}
= \sum_i w_{i,j}\alpha_{i,k'}\alpha_{i,k''}$ \\

\addlinespace

$\mathbf{C}$
& $\displaystyle \frac{\partial L}{\partial c_{j,k}}
= \sum_{i,\ell} g_{i,\ell}\beta_{i,j,\ell,k}$
& $\displaystyle h^{(k)}_{j',j''}
= \sum_{i,\ell} w_{i,\ell}
\beta_{i,j',\ell,k}
\beta_{i,j'',\ell,k}$ \\

\bottomrule
\end{tabular}

\vspace{1ex}

\footnotesize
For the Multinomial likelihood, use
$\alpha_{i,k,p} = \sum_m p_{i,m,p}c_{m,k}$ and
$\beta_{i,j,j',k,p} = p_{i,j,p}s_{k,j'}$.
The Hessian terms are used by the quadratic-approximation updates; the likelihood-based PCHA baseline uses only the corresponding gradients.

\end{table}
\subsection{Predictive Likelihood}
\label{sec:predictive_likelihood}

Let $\mathbf{X} \in \mathbb{R}^{M \times N}$ denote a dataset with $M$ features and $N$ samples. 
For a fixed number of archetypes $K$, the model is fitted on training data 
$\mathbf{X}_{\mathrm{train}} \in \mathbb{R}^{M \times N_{\mathrm{train}}}$, providing coefficient matrices 
$\mathbf{C}$ and $\mathbf{S}$. The fitted archetypes and reconstruction are respectively given by
\begin{equation}
    \mathbf{Z} = \mathbf{X}^{\mathrm{train}}\mathbf{C},    \qquad
    \mathbf{R}^{\mathrm{train}} = \mathbf{Z}\mathbf{S},
\end{equation}
where the columns $\mathbf{r}_j = (\mathbf{R}^{\mathrm{train}})_{\cdot j}$ define the reconstruction associated with the $j^{th}$ training sample.

To compare different values of $K$, we evaluate how well the fitted archetypal representation $\mathbf{Z}$ based on $\mathbf{X}^{\mathbf{train}}$ predicts held-out observations. 
For a test observation $\mathbf{x}$, we approximate the predictive distribution by the empirical mixture over the reconstructed training  parameters,
\begin{equation}
\begin{aligned}
p(\mathbf{x}^{\mathrm{test}} \mid \mathbf{X}^{\mathrm{train}}, \mathbf{Z})
&= \int 
p(\mathbf{x}^{\mathrm{test}} \mid \mathbf{Z}, \mathbf{s})
p(\mathbf{s} \mid \mathbf{X}^{\mathrm{train}}, \mathbf{Z})
\, d\mathbf{s} \\
&\approx \frac{1}{N_{\mathrm{train}}}
\sum_{j=1}^{N_{\mathrm{train}}}
p(\mathbf{x}^{\mathrm{test}} \mid \mathbf{r}^{\mathrm{train}}_j).
\end{aligned}
\label{eq:predictive_mixture}
\end{equation}
The posterior is according to Bayes' theorem given by 
\begin{equation}
p(\mathbf{s} \mid \mathbf{X}^{\mathrm{train}}, \mathbf{Z})=\frac{p(\mathbf{X}^{\mathrm{train}} | \mathbf{Z},\mathbf{s})p(\mathbf{s})}{\int p(\mathbf{X}^{\mathrm{train}} | \mathbf{Z},\mathbf{s})p(\mathbf{s})d\mathbf{s}}.
\end{equation}
Whereas $p(\mathbf{X}^{\mathrm{train}} | \mathbf{Z},\mathbf{s})$ is the likelihood, the prior $p(\mathbf{s})$ can be any distribution with support over the probability simplex such as the Dirichlet distribution or continuous categorical distribution \cite{CaberoFayos2020FindingQuestionnaires}.
However, the marginalization over $\mathbf{s}$ is in general intractable preventing the direct evaluation of the predictive likelihood. By imposing a uniform prior over the simplex corresponding to $p(\mathbf{s})= \text{Dirichlet}(\mathbf{s}|\mathbf{1})$ the shape of the posterior distribution is only influenced by the likelihood. Consequently, $p(\mathbf{s}\mid \mathbf{X}^{\mathrm{train}},\mathbf{Z})\propto p(\mathbf{X}^{\mathrm{train}}\mid \mathbf{Z},\mathbf{s})$ and by using the training samples to approximate the integral assuming that the learned $\mathbf{S}^\text{train}$ are drawn unbiased from the posterior distribution 
$\mathbf{s}^{\mathrm{train}}_j\sim p(\mathbf{s}\mid \mathbf{X}^{\mathrm{train}},\mathbf{Z})$ under this uniform prior we obtain when using $\mathbf{r}^{\mathrm{train}}_j=\mathbf{Z}\mathbf{s}^{\mathrm{train}}_j$ the approximation of the predictive likelihood in eq. \eqref{eq:predictive_mixture}. Thus, for a test set 
$\mathbf{X}_{\mathrm{test}} = \{x_i\}_{i=1}^{N_{\mathrm{test}}}$, the predictive log-likelihood is
\begin{equation}
    \mathcal{L}_{\mathrm{pred}}
    =
    \sum_{i=1}^{N_{\mathrm{test}}}
    \log
    \left[
    \frac{1}{N_{\mathrm{train}}}
    \sum_{j=1}^{N_{\mathrm{train}}}
    p(x_i \mid r_j)
    \right].
    \label{eq:predictive_loglik}
\end{equation}
For numerical stability, each term is evaluated using
\begin{equation}
    \log p(\mathbf{x}_i \mid \mathbf{r}_j)
    =
    \operatorname{logsumexp}_{j}
    \left\{
    \log p(\mathbf{x}_i \mid \mathbf{r}_j)
    \right\}
    -
    \log N_{\mathrm{train}} .
\end{equation}

The likelihood contribution $\log p(\mathbf{x} \mid \mathbf{r}_j)$ depends on the observation model. 
The models used in this work are summarized in Table~\ref{tab:predictive_likelihoods}. 
Terms independent of $\mathbf{r}_j$ are omitted for the Poisson and multinomial models, since they do not affect model comparison across values of $K$.

\begin{table}[t]
\centering
\caption{Log-likelihood contributions used in the predictive likelihood criterion.}
\label{tab:predictive_likelihoods}
\begin{tabular}{ll}
\hline
Model & $\log p(\mathbf{x} \mid \mathbf{r}_j)$ \\
\hline
Bernoulli 
&
$\displaystyle
\sum_{f=1}^{M}
\left[
x_m \log r_{mj}
+
(1-x_m)\log(1-r_{mj})
\right]$
\\[2.0ex]

Poisson 
&
$\displaystyle
\sum_{m=1}^{M}
\left[
x_m \log r_{mj}
-
r_{jm}
\right]$
\\[2.0ex]

Gaussian 
&
$\displaystyle
-\frac{1}{2}
\left(
\frac{\|\mathbf{x}-\mathbf{r}_j\|_2^2}{\sigma^2}
+
d\log(2\pi)
+
d\log\sigma^2
\right)$
\\[2.0ex]

Multinomial 
&
$\displaystyle
\sum_{m=1}^{M}
x_m \log r_{mj}$
\\[1.5ex]
\hline
\end{tabular}
\end{table}

For the Gaussian model, we further need to specify the isotropic covariance 
$\mathbf{x} \sim \mathcal{N}(\mathbf{r}_j,\sigma^2 \mathbf{I}_M)$, which we estimate by maximum likelihood from the training residuals as
\begin{equation}
    \sigma^2
    =
    \frac{1}{MN_{\mathrm{train}}}
    \left\|
    \mathbf{X}_{\mathrm{train}}
    -
    \mathbf{R}_{\mathrm{train}}
    \right\|_F^2 .
\end{equation}

In $F$-fold cross-validation, the predictive score for each candidate number of archetypes is obtained by summing the held-out predictive log-likelihoods across folds,
\begin{equation}
    \mathrm{CV}(K)
    =
    \sum_{f=1}^{F}
    \mathcal{L}_{\mathrm{pred}}^{(f)}(K),
    \quad
    K^\star
    =
    \arg\max_K \mathrm{CV}(K).
\end{equation}

\subsection{Normalized Mutual Information}
Normalized Mutual Information (NMI) is used together with the predictive likelihood to evaluate the solution. This was first introduced to AA in \cite{Hinrich2016ArchetypalData}. The method evaluates the columns in $\mathbf{S}$ as probability distributions and computes the consistency across different initializations of $\mathbf{S}$. A high NMI indicates a stable solution whereas a low NMI indicates an unstable solution. The NMI between two solutions $\mathbf{S}^{r}$ and $\mathbf{S}^{r'}$ is given by: 
\begin{equation}
    \operatorname{NMI}(\mathbf{S}^{r}, \mathbf{S}^{r'})
    =
    \frac{
    2\operatorname{MI}(\mathbf{S}^{r}, \mathbf{S}^{r'})
    }{
    \operatorname{H}(\mathbf{S}^{r})+\operatorname{H}(\mathbf{S}^{r'})
    },
\end{equation}

where the mutual information is defined as

\begin{equation}
    \operatorname{MI}(\mathbf{S}^{r}, \mathbf{S}^{r'})
    =
    \sum_{k=1}^{K}\sum_{k'=1}^{K}
    p(k,k')
    \log
    \frac{
    p(k,k')
    }{
    p(k)p(k')
    },
\end{equation}

with

\begin{equation}
\begin{aligned}
    p(k,k')
    &=
    \frac{1}{N}
    \sum_{n=1}^{N}
    s^{r}_{k,n}s^{r'}_{k',n}, \\
    p(k)
    &=
    \frac{1}{N}
    \sum_{n=1}^{N}
    s^{r}_{k,n}, \quad 
    p(k')
    =
    \frac{1}{N}
    \sum_{n=1}^{N}
    s^{r'}_{k',n}.
\end{aligned}
\end{equation}

The entropy is given by

\begin{equation}
    \operatorname{H}(\mathbf{S}^{r})
    =
    -\sum_{k=1}^{K} p(k)\log p(k).
\end{equation}

\subsection{Computational complexity and memory usage.}

The alternating algorithm consists of an $\mathbf{S}$-update and a $\mathbf{C}$-update. In the $\mathbf{S}$-update, each column $\mathbf{s}_j$ lies in the $K$-dimensional simplex and is updated using pairwise SMO steps. A full sweep over all archetype pairs costs $\mathcal{O}(NK^2)$ once the likelihood-specific gradients and Hessian terms have been formed. Constructing these quantities typically costs $\mathcal{O}(MNK)$. The main scalability challenge occurs in the $\mathbf{C}$-update, since each column $\mathbf{c}_k$ lies in an $N$-dimensional simplex. A na\"{\i}ve active-set method may require storing and solving systems involving large subsets of the $N$ observations, leading to memory costs as high as $\mathcal{O}(N^2)$ if the full Hessian is materialized. We avoid this by imposing a maximum active-set size $N_{\max}$. The reduced solve then involves at most $N_{\max}$ explicit active variables, giving a per-solve cost of $\mathcal{O}(N_{\max}^3)$ and reduced Hessian storage of $\mathcal{O}(N_{\max}^2)$. When the active set reaches $N_{\max}$, its current contribution is condensed into a proxy vector. This proxy preserves the accumulated convex combination while resetting the explicit active set. Consequently, the optimization can continue to explore new active observations without allowing the reduced system to grow with $N$. The max-active mechanism therefore changes the practical memory requirement of the $\mathbf{C}$-update from being governed by the full sample size $N$ to being governed primarily by $N_{\max}$, while still representing the full coefficient vector through the combination of explicit active entries and the proxy component.

\subsection{Extension of PCHA to discrete data}
In addition to introducing a new optimization scheme, we  extended the PCHA algorithm \cite{Mrup2012ArchetypalMining} to discrete data by deriving likelihood-consistent gradient expressions for Bernoulli, Poisson and Multinomial distributed observations. This extension enables principled probabilistic modeling within the established PCHA architecture and facilitates direct empirical comparison with this procedure in the experimentation (see gradient terms in tables \ref{tab:likelihood_factors} and \ref{tab:generic_update_components}. for specific $\mathbf{C}$ and $\mathbf{S}$ steps).

\section{Results}
We first present the optimization behavior analyzed on synthetic data, where the true number of archetypes is known and subsequently on three real datasets.

\subsection{Reproducibility}
All experiments were run from the DTU hpc (The Technical University of Denmark's high performance computer) queue as single-core CPU jobs, with identical resource requests within each experiment. 
The implementation used in this paper is available in the Distribution Agnostic
Archetypal Analysis repository \cite{wedenborg2026daaa_software}.

Unless stated otherwise, models were evaluated over five random seeds ($0-4$) using 5-fold predictive cross-validation. Synthetic experiments used data seed 0, $K^\star = 5$, candidate $K \in \{2,3,4,5,6,8,10,15\}$, DAAA used pairwise two-coordinate SMO the active-set cap was set to $25\%$ unless explicitly stated. 
For direct method comparisons, fold splits and the initial values of $\mathbf{C}$ and $\mathbf{S}$ were controlled by using the same random seeds across methods. All methods were initialized using five seeds, ($0-4$). The initial matrices were generated from uniformly sampled random values, transformed using the absolute logarithm, and subsequently column-normalized.

For DAAA and PCHA, additional sparsity was imposed on $\mathbf{C}$ by setting all entries below $0.5$ to zero. This sparsification step was omitted for PAA \cite{Seth2016ProbabilisticAnalysis}, as PAA was substantially more sensitive to initialization and did not perform reliably under this sparse initialization.

\subsection{Synthetic Data Experiments}
 The aim of the synthetic data experimentation was to analyze the feasibility and recoverability of the proposed framework. All synthetic data sets were generated with $K=5$ archetypes and were fitted with candidate values $K \in \{2,3,4,5,6,8,10,15\}$. For each sample $n$, mixing weights were drawn from a Dirichlet distribution, $h_n \sim \operatorname{Dirichlet}(\alpha \mathbf{1}_{K^*}),  \alpha = 0.3,$ such that the observations were concentrated near a small number of archetypes. for the Gaussian, Bernoulli, and Poisson experiments, archetype profiles were drawn as sparse binary vectors, $\eta_{mk} \sim \operatorname{Bernoulli}(s), s = 0.3,$ with duplicate archetypes rejected. Conditional on these archetypes and the sample weights, observations were generated as, $x_{m,n} = \eta_m^\top h_n + \epsilon_{m,n},
     \epsilon_{m,n} \sim \mathcal{N}(0,\sigma^2), \sigma = 0.1,$
for the Gaussian case. $x_{m,n} \sim \operatorname{Bernoulli}\!\left(\min\{\eta_m^\top h_n,0.99\}\right)$ for the Bernoulli case and $x_{m,n} \sim \operatorname{Poisson}\!\left(r\,\eta_m^\top h_n\right), r = 5,$ for Poisson. For the multinomial experiment, each archetype-feature
pair was instead assigned a categorical probability vector, $\eta_{mk} \sim \operatorname{Dirichlet}(s\mathbf{1}_{C}), C = 10,$ and the sample-specific category probabilities were formed by convex mixing, $p_{m,n} = \sum_{k=1}^{K_0} h_{n,k}\eta_{m,k}.$ The observed categorical vector was then drawn from a one-trial Multinomial
distribution, $x_{mj} \sim \operatorname{Multinomial}(1,p_{mj})$. All experiments used $N = 1000$ samples and $M=500$ features.

\begin{figure}
    \centering
    \includegraphics[width=\linewidth]{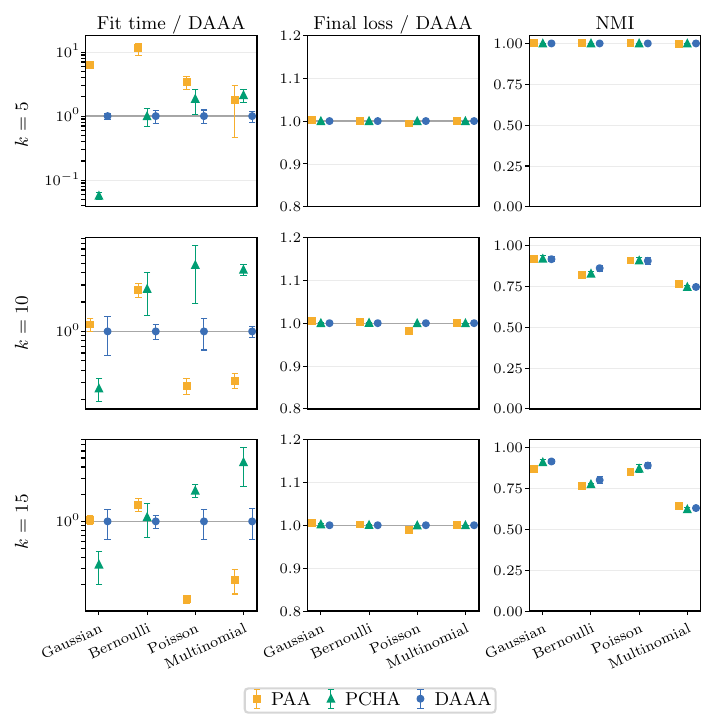}
    \caption{\textbf{Cross-distribution robustness of synthetic data}, for $K \in \{5, 10, 15\}$, with $K^\star=5$. Each point shows the average performance of PAA, PCHA, and DAAA across repeated initializations for a fixed synthetic data setting. Runtime (seconds) and final loss (NLL) are reported relative to DAAA within each distribution, so values above one indicate slower runtime or higher loss than DAAA. NMI measures stability across random initializations, with larger values indicating more consistent recovered archetypal representations.
    }
    \label{fig:SynthExperiment}
\end{figure}

In this simple setting, all methods trivially recovered $K^{\star}$. 
From \autoref{fig:SynthExperiment}, we observe the performance of PAA, PCHA, and DAAA relative to DAAA across four data-generating distributions when specifying the number of estimated components to $k \in {5, 10, 15}$. All three methods reach essentially the same final loss in every setting: the loss ratio remains within $\pm 2\%$, confirming that the choice of algorithm does not compromise solution quality. 

The computational picture is more varied. At $K = 5$, PAA is substantially slower than DAAA especially for Gaussian and Bernoulli data. PCHA is dramatically faster on Gaussian data but slower for Poisson and Multinomial likelihoods. This advantage for PCHA on Gaussian data narrows as $K$ increases; at $K = 10$ and $K = 15$, PCHA is up to $4\times$ slower than DAAA on Multinomial data. PAA's relative cost also decreases with $K$: by $K = 15$ it is comparable to, or faster than, DAAA on Poisson and Multinomial data, while remaining slower on Gaussian and Bernoulli data. DAAA thus occupies a competitive position across all distributions and archetype counts.

Optimization stability, measured as the pairwise NMI between solutions from different initializations on the same dataset declines with $K$ for all methods. At $K = 5$ stability is near-perfect ($\text{NMI} \approx 1$) across all distributions, whereas at $K = 15$ it drops to approximately 0.85–0.90 for Gaussian data and as low as 0.60–0.65 for Multinomial data. All three methods exhibit comparable NMI at every setting, indicating that the decreasing stability reflects the geometry of the problem ($K^\star = 5$) rather than differences between optimizers.

\begin{figure}
    \centering
    \includegraphics[width=\linewidth]{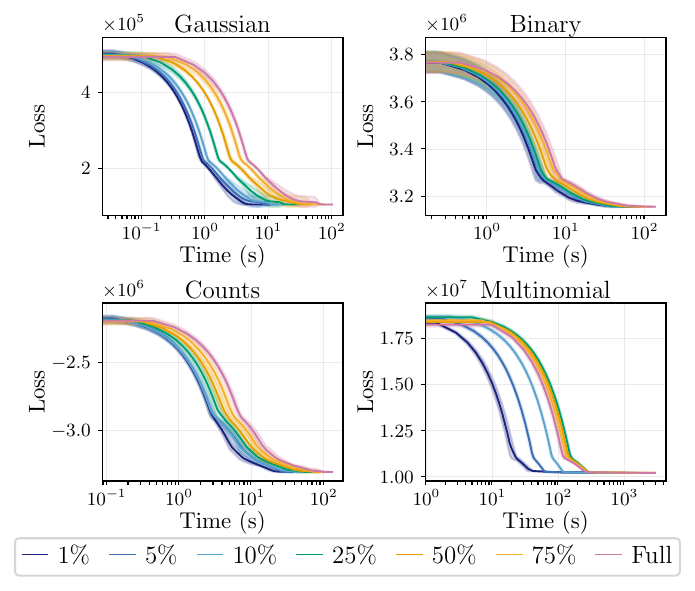}
    \caption{\textbf{Convergence of Archetypal Analysis under varying active set sizes}. Training loss as a function of wall-clock time for active-set fractions $N_\text{max} \in \{ 1\%, 5\%, 10\%, 25\%, 50\%, 75\%, 100\% \}$. Where $100\%$ corresponds to using all $N$ samples.
    Shaded bands show mean $\pm$ std across five random seeds. The x-axis is on log scale.}
    \label{fig:n_max_timings}
\end{figure}
To test the feasibility of the constrained active set we examined the loss and convergence speed relative to the size of the active set in \autoref{fig:n_max_timings}. Here we see that smaller active sets achieve faster early descent across all distributions. For Gaussian, Binary, and Counts data, fractions $\geq 5\%$ converge to a final loss indistinguishable from the full active set. For Multinomial data, fractions $\geq 25\%$ suffice, while $1\%$ and $5\%$ incur a persistent gap. 

\begin{figure}
    \centering
    \includegraphics[width=\linewidth]{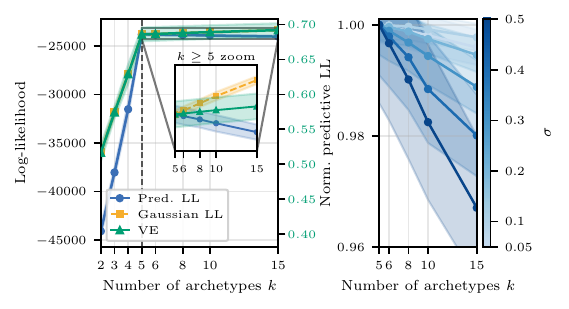}
    \caption{\textbf{Model selection for Gaussian archetypal analysis on synthetic data} ($K = 5$ true archetypes). \textit{Left}: Cross-validated predictive log-likelihood (Pred. LL), Gaussian training log-likelihood (Gaussian LL), and variance explained (VE) as functions of $K$ at a single noise level. The predictive LL peaks at the true $K^* = 5$ (dashed line), while Gaussian LL and VE increase monotonically, failing to identify the correct model complexity (inset). \textit{Right}: Predictive LL normalized by its maximum across noise levels $\sigma \in \{0.05, 0.1, 0.2, 0.3, 0.4, 0.5\}$; shaded bands show variance across seeds. The predictive LL is maximised at $K^\star = 5$ across all noise levels, and the penalty for overspecifying $K$ grows with $\sigma$, reflecting greater sensitivity to overfitting under high noise.}
    \label{fig:synth_predLL}
\end{figure}

The proposed predictive log-likelihood framework for determining the optimal $K$ was tested in a Gaussian setting for simplicity (noise is easy to model and control through the variance $\sigma$). In \autoref{fig:synth_predLL} it can be seen that while the Variance Explained and NMI gradually increases after the optimal $K$ the predictive log-likelihood declines, showing a usable framework for quantitative analysis of $K$. This signal becomes clearer as more noise is added to the data generation process.

\subsection{Real Data experiments}

\begin{figure}
    \centering
    \includegraphics[width=\linewidth]{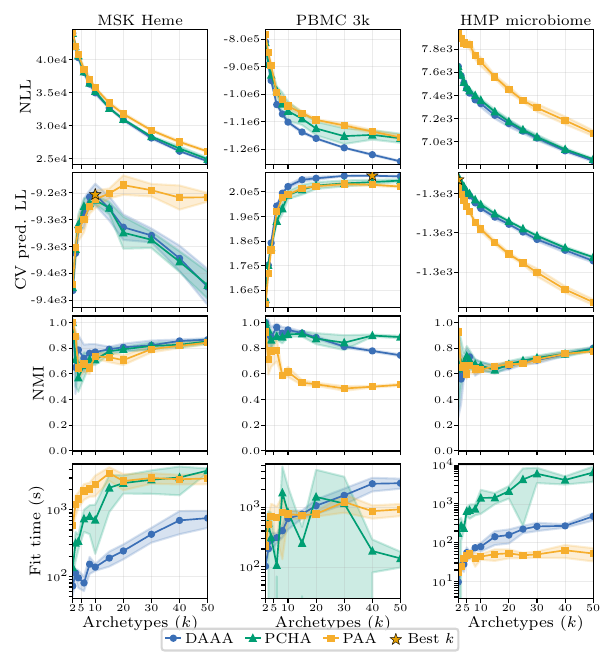}
    \caption{\textbf{Real-data benchmarks across three datasets and distributions}. Each column corresponds to one dataset (MSK Heme, Bernoulli; PBMC 3k, Poisson; HMP microbiome, Multinomial) and each row to one metric: training loss (NLL), cross-validated predictive log-likelihood (star marks the DAAA selected $K$), stability NMI across random initialisations, and wall-clock fit time. DAAA uses \texttt{max\_active} = 25\% of samples. PCHA and PAA use the full dataset. Shaded bands denote $\pm 1$ standard deviation across five random seeds.}
    \label{fig:real_data_stats}
\end{figure}

Having established the behavior of the method on synthetic data, we next evaluate it on three real data sets chosen for the three discrete data based likelihoods implemented in the framework. The purpose of these experiments is twofold. We first ask whether the likelihood-based model objective gives competitive fits, stable solutions, and useful predictive likelihoods on data sets whose structure is clearly non-Gaussian.  We then inspect whether the resulting archetypes are interpretable within their domain. We consider the following three domains; binary data based on blood cancer mutations, count data based on gene programs and cell types for single-cell RNA sequencing, and categorical data based on microbiome profiles
across body sites, these datasets are described below.

The binary Bernoulli likelihood based experiment uses the MSK-IMPACT Heme cohort\cite{ptashkin_impact_heme_2023}, a targeted sequencing dataset of hematologic malignancies. We construct a binary sample-by-gene somatic mutation matrix from the single nucleotide Variant (SNV) or insertion/deletion (indel), i.e. SNV/indel calls, setting an entry to one if a sample contains at least one somatic SNV or indel in the corresponding gene and zero otherwise.

\begin{figure}
    \centering
    \includegraphics[width=\linewidth]{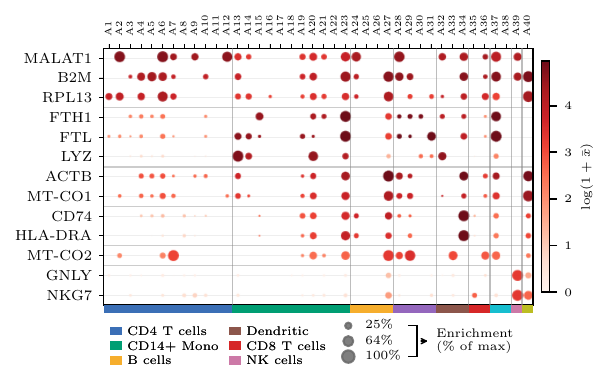}
    \caption{\textbf{Within cell type gene groups across $K = 40$ archetypes (PBMC3k scRNA-seq)}
    Dot plot of gene expression across 40 archetypes fitted by DAAA to the PBMC3k dataset (2,700 peripheral blood mononuclear cells, 2,000 highly variable genes. Archetypes (A1–A40, columns) are grouped by dominant cell type (coloured bar) and sorted by decreasing cell-type purity within each group. Genes (rows) are selected as the top drivers of within-cell-type expression variance across sub-archetypes. Dot size (square-root scaled, normalised per gene) encodes within-cell-type enrichment, the archetype mean expression minus the mean across all archetypes of the same cell type, isolating sub-state variation from the shared cell-type background. Dot colour encodes absolute expression as log(1 + mean raw count), compressing the dynamic range of integer counts. CD4 T cells (A1–A12) resolve into 12 sub-archetypes spanning a translational-activity gradient (MALAT1, RPL13), an MHC-I axis (B2M), and a mitochondrial-high state (MT-CO1); CD14$^+$ monocytes (A13–A23) separate into LYZ-high classical and FTL/FTH1-enriched sub-states}
    \label{fig:PBMCK3}
\end{figure}

The count Poisson likelihood based experiment uses the PBMC3k experiment uses the 10x Genomics peripheral blood mononuclear cell data set available through \texttt{scanpy}\cite{10x_pbmc3k_2016,Wolf2018SCANPY}.  After standard filtering, we retain the $m=2000$ highly variable genes while fitting the model to the original integer count matrix.  This data set is therefore treated as count-valued and modeled with the Poisson likelihood.  The fitted archetypes are interpreted as extreme gene-expression profiles, and the sample memberships are compared with known PBMC cell-type annotations in the downstream diagnostic plot in \autoref{fig:PBMCK3}.

\begin{figure}
    \centering
    \includegraphics[width=\linewidth]{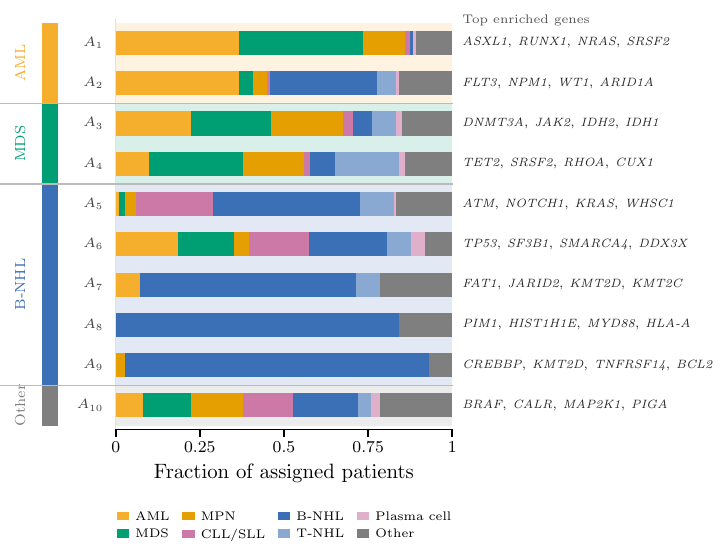}
    \caption{\textbf{Interpretability of learned archetypes on the MSK-IMPACT Heme somatic mutation cohort}. Each row corresponds to one of $K=10$ archetypes. Bar colors show the fraction of hard-assigned patients from each cancer type group. Archetypes are sorted by dominant lineage, indicated by the colored strip and vertical label on the left. The four most enriched mutations per archetype, the largest positive deviations from the dataset, wide mean mutation rate, are listed to the right. Without using cancer type labels during fitting, the method recovers clinically coherent structure: the two AML archetypes are defined by chromatin-regulator mutations (ASXL1, RUNX1) and the canonical NPM1/FLT3 driver pair; the MDS archetypes are characterised by DNA-methylation genes (DNMT3A, TET2, IDH1/2); and five distinct B-NHL archetypes separate known molecular subtypes, including an ABC-DLBCL profile (MYD88, CD79B) and a follicular lymphoma profile (BCL2, EZH2, CREBBP). Archetype $A_{10}$ is a diffuse, mixed-lineage profile with no single dominant mutation signal, reflecting the shared background mutation burden across all subtypes.}
    \label{fig:binary}
\end{figure}

The categorical multinomial likelihood based experiment uses a Human Microbiome Project Phase 1 16S tensor aggregated at genus level \cite{hmp_consortium_2012}. The tensor has taxa, subjects, and body sites as its
three axes; in the checked-in preprocessing report this corresponds to $589$ taxa, $235$ subjects, and $18$ body sites.  Since each subject-taxon slice describes a composition over body sites, this data set is fitted with the multinomial likelihood.  The resulting archetypes are intended to capture recurrent microbial profiles whose mass is concentrated on different anatomical sites or combinations of sites.

\begin{figure}
    \centering
    \includegraphics[width=\linewidth]{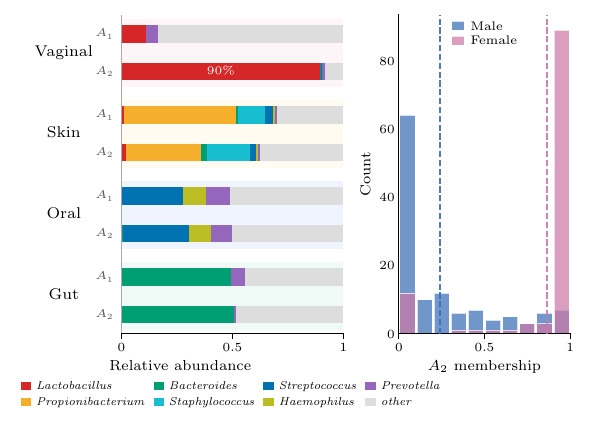}
    \caption{\textbf{Multinomial Archetype microbiome profiles across four anatomical body-site groups} (235 subjects, 18 body sites, 589 genera). (\textit{Left}) Each bar shows the mean relative abundance of the dominant genera, averaged across all body sites within the group (Gut: 1 site; Oral: 9 sites; Skin: 5 sites; Vaginal: 3 sites). Archetype profiles are computed as the anchor-weighted mean composition $\sum_i C_{ik} X_{ijp}$, then averaged over sites within each group. A1 and A2 are nearly indistinguishable at gut and oral sites. Skin shows a moderate enrichment of Propionibacterium in A1 and Staphylococcus in A2. Vaginal sites show the main difference: A2 is 90\% Lactobacillus on average, compared to 11\% in A1. (\textit{Right}) Histogram of A2 membership weights by sex (dashed lines: group means). Male subjects cluster near zero (mean 0.24) and female subjects near one (mean 0.86), demonstrating that the model unsupervised recovers sex as the dominant axis of variation through the vaginal Lactobacillus signal.}
    \label{fig:multinomial}
\end{figure}

For each data set, we fit DAAA, PAA, and PCHA over a grid of candidate numbers of archetypes ($2-50$) and random initialization seeds ($0-4$). We report the final training objective, wall-clock fitting time measured in seconds, and pairwise soft NMI between membership matrices across seeds as a measure of optimization stability. We also compute five-fold predictive log-likelihood. We use this as primary criterion to select the number of archetypes as the real data do not come with known ground-truth values of $K$.

From \autoref{fig:real_data_stats} we observe that DAAA compared to PAA generally produced more stable solutions and better likelihood fits, suggesting that the proposed optimization strategy is better suited for discrete real-world data, while the fit time was generally more favorable for DAAA than PCHA both having similar solution quality. From the predictive log-likelihood plots we saw that model performance were similar for the DAAA and PCHA approaches however, we observe discrepancy to PAA in particular for the MSK Heme mutation data where overfitting was less of an issue for PAA. From the figure we further observe that across the three real-world datasets, the proposed DAAA framework selected different model complexities that reflected the structure of each domain: (K=10) for the MSK Heme mutation data, (K=40) for the PBMC3k single-cell RNA-seq data, and (K=2) for the HMP microbiome data. In each case, the predictive likelihood criterion identified archetypes that were biologically interpretable, including cancer-type-specific mutation profiles as illustrated in Figure \ref{fig:binary}. These profiles (archetypes) split into three distinct main categories; one driven by AML, one by MDS and one B-NHL. These correspond to myeloid and lymphoid types of blood cancer, with AML and MDS representing myeloid malignancies and B-NHL representing a lymphoid malignancy. Distinct expression patterns across gene groups with archetypes primarily described by certain PBMC cell-types as illustrated in Figure \ref{fig:PBMCK3}, and microbiome variation driven largely by body-site and sex-associated composition as illustrated in Figure \ref{fig:multinomial}.

\section{Discussion}

The synthetic study demonstrated similar performance across methods for AA inference for the different distributions (\autoref{fig:SynthExperiment}), notably no methods were best across all the synthetic generated data in terms of fit time. DAAA worked in general more favorably for discrete data than the PCHA. However, PAA was more computationally efficient for the Poisson and Multinomial when the model was overparameterized ($K>K^\star$). 

Notably, the constrained active set procedure enhanced convergence with minimal compromise in solution quality, supporting the utility of the proposed active set scaling. Although this scaling was evaluated here in the context of AA, we expect that similar gains may be achievable in other modeling regimes that rely on active set procedures.

For the real data, we observe poor performance from the PAA method across all regimes when compared to both the DAAA and PCHA in terms of solution quality (NLL) and stability (NMI). Whereas PCHA and DAAA here had similar performance the PCHA was notably slower than the proposed DAAA. Again, we observe that while the PAA was slower for Bernoulli, it had favorable fit time for the multinomial data. For Poisson the PAA only had favorable fit time in higher model order regimes. The poorer log-likelihood obtained by PAA should most naturally be interpreted as an optimization issue, rather than as evidence against the archetypal representation itself. 
Unlike PCHA and DAAA, PAA relies on multiplicative updates. Such updates preserve positivity and are monotone under suitable conditions, but they suffer from several limitations, small coefficients tend to remain small, mass is reallocated only gradually, and convergence can be slow when the likelihood is ill-conditioned or close to the boundary of the parameter space. This is particularly consequential for Bernoulli, Poisson, and multinomial likelihoods, where underestimating rare events, zero counts, or low-probability categories can have a large effect on the log-likelihood. Thus, PAA may capture the dominant archetypal structure while still producing less well-calibrated fitted probabilities or intensities, leading to substantially lower likelihood values than PCHA and DAAA.

Importantly, we proposed generalization as estimated based on our predictive log-likelihood framework as a way to tackle the open problem of determining the number of archetypes. For the synthetic data, we observe that the method correctly identifies the model order with the peak in generalization error enhanced as the noise level increased and where overparameterization should be more influenced by noise, see \autoref{fig:synth_predLL}. This points to the validity and practical use of the predictive likelihood framework for model order estimation in AA. Notably for the real data we also observed peaks in the generalization error indicating a suitable model order for the three datasets respectively modeled using the Bernoulli, Poisson, and multinomial likelihoods.
When inspecting the associated identified models by the DAAA, as indicated by a star in \autoref{fig:real_data_stats}, we observed meaningful extracted representations in which the AA procedures extracted distinct patterns of variation in the datasets. 

Whereas we presently focus on the modeling of discrete data we note that the proposed optimization- and generalization framework for estimating the AA model and model order respectively, can be used beyond the considered likelihoods. 

\subsection{Limitations}
While the proposed DAAA framework exhibits stable solutions across archetypes, distributions and datasets AA remains inherently non-convex and initialization sensitive. Similarly the method depends on data following a specific likelihood formulation and focuses only on separable likelihoods. This framework allows for the extension to heterogeneous likelihood formulations, where each data feature can have different distributions, however this is left for future work. 
In this paper, we introduced a constrained active set size to improve scalability. From \autoref{fig:n_max_timings} this proved to a be feasible and competitive solution for AA that can favorably trade exactness for scalability, especially if the limit to the active set is very restrictive. As such, we observed favorable fit time when comparing to PCHA while achieving similar solution qualities. However the constrained active set comes with a cost in numeric precision and should not be used when such numeric solution precision is highly critical. 

Although the predictive log-likelihood gave clear indications about the optimal $K$, this was only qualitatively assessed. The CV predictive log-likelihood is the most principled way of choosing $K$ for AA and well established in other statistical domains but the resulting archetypes are only validated qualitatively. A more principled analysis across many domains and datasets we leave for future work.
Another limitation of the predictive likelihood criterion arises in limited-data settings. The predictive score is computed by using the fitted training representation, in particular the training $\mathbf{S}$-matrix, as an empirical approximation to the latent mixing distribution. When the number of training samples is small, this empirical distribution can be sparse and unstable, individual training samples can carry disproportionate weight, and the set of observed archetypal mixtures may not adequately represent the mixtures present in the held-out data. As a result the predictive likelihood can become sensitive to the particular train-test split and may penalize otherwise reasonable archetypes simply because the training $\mathbf{S}$-matrix provides a poor finite-sample approximation to the underlying population distribution. This suggests that predictive likelihood is most reliable when the training set is sufficiently large to capture the relevant variation in mixture proportions, and that additional smoothing or hierarchical modeling of the latent mixing distribution may be beneficial in low-sample regimes.
\section{Conclusion}
We introduced an efficient and generalizable likelihood-based framework for AA for discrete data. By combining local quadratic likelihood approximations with SMO updates for the mixture weights and constrained active-set updates for the archetype construction weights, the method provides an efficient optimization scheme as highlighted for the considered Bernoulli, Poisson, and multinomial observation models. 

We further derived a predictive likelihood criterion for selecting the number of archetypes. In synthetic experiments, this criterion recovered the true model complexity, whereas the NLL training reconstruction-based measures tended to favor larger models with no clear indication of a suitable number of components. The real-data examples further showed that the optimally defined number of archetypes based on generalization using the proposed predictive likelihood framework could be interpreted as meaningful domain-specific profiles characterizing cancer mutations, PBMC gene-expression programs, and microbiome body-site variation.

Overall, the proposed method offers an efficient alternative approach to likelihood based inference in AA with the proposed predictive likelihood framework providing a simple to use model order assessment procedure that enables model assessment in terms of how well the extracted AA polytope generalizable to new data. 

\bibliography{references}


\section{Biography Section}
\begin{IEEEbiography}[{\includegraphics[width=1in,height=1.25in,clip,keepaspectratio]{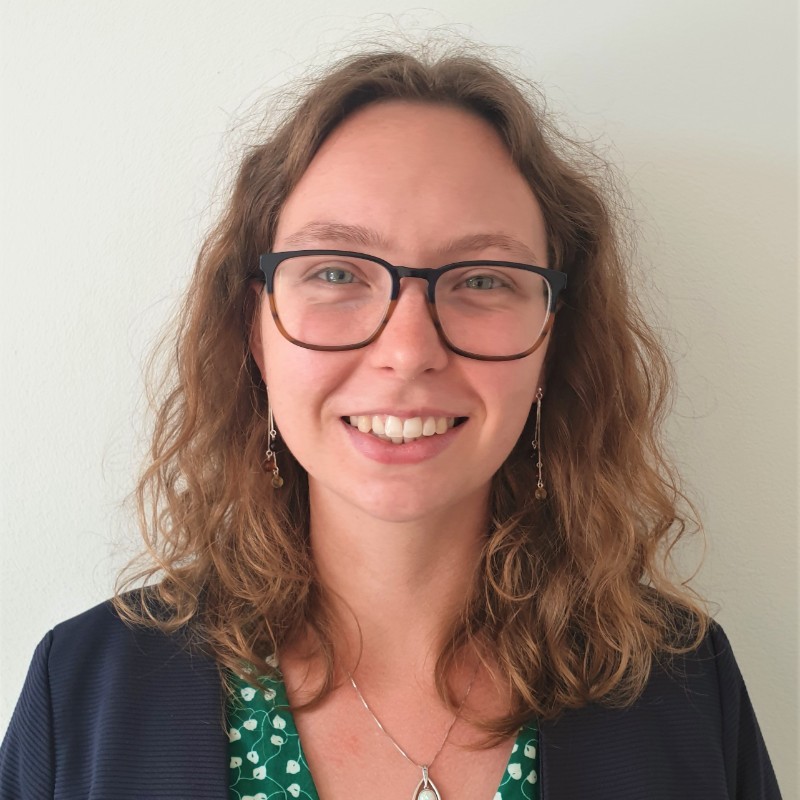}}]{A. Emilie J. Wedenborg}
has a B.Sc. in Quantitative Biology and Disease Modeling and a M.Sc. in mathematical modeling and computing both from the Technical University of Denmark (DTU). She is currently a PhD. fellow at DTU Compute, section of Cognitive Systems. Her research interests include interpretable modeling approaches, especially related to life-sciences. 
\end{IEEEbiography}

\begin{IEEEbiography}[{\includegraphics[width=1in,height=1.25in,clip,keepaspectratio]{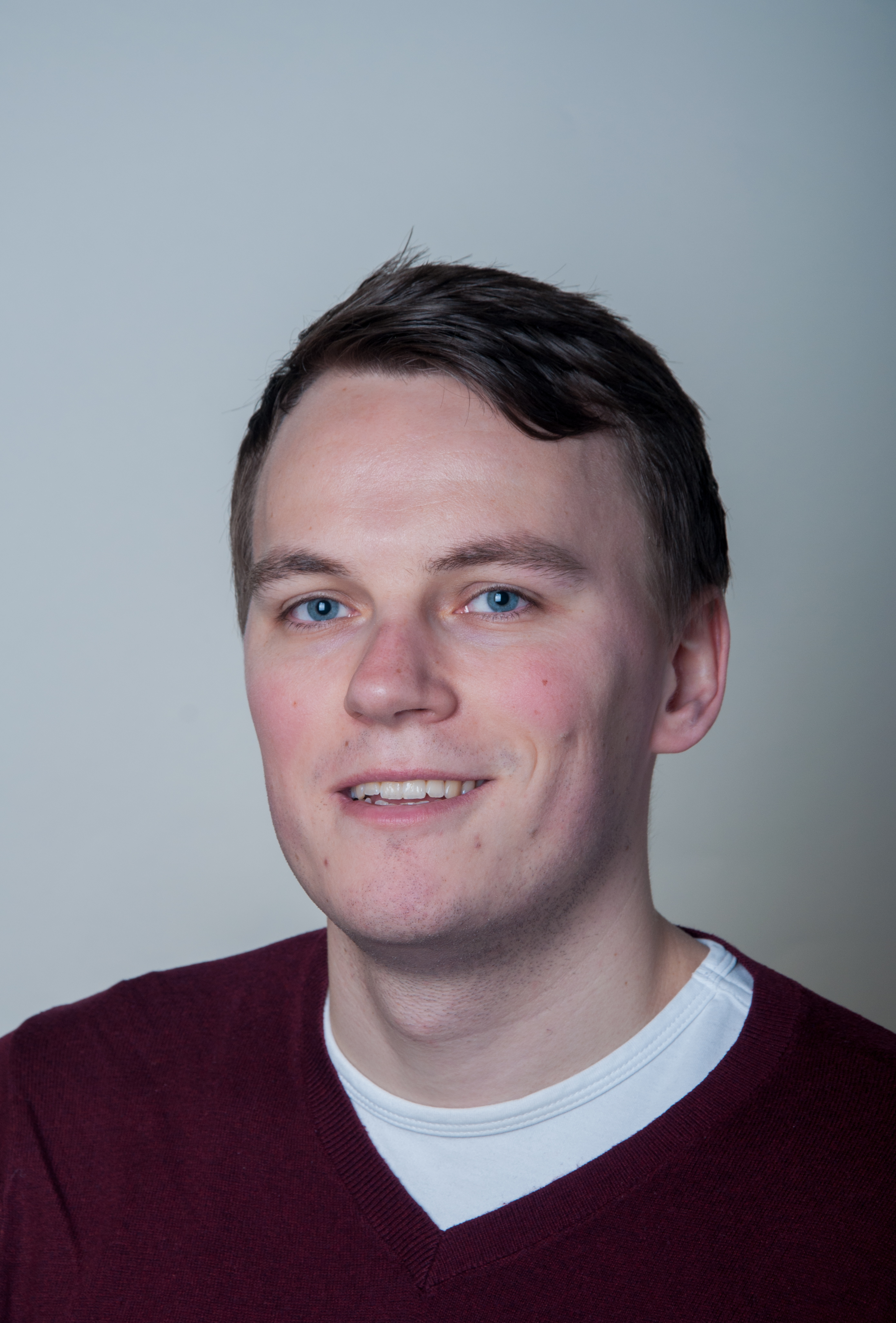}}]{Jesper Løve Hinrich}
received his M.Sc. Eng. and PhD degrees in mathematical modelling and computation from the Technical University of Denmark. He is currently a postdoc at DTU Compute, Section for Cognitive Systems. His research focuses on tensor networks and statistical machine learning, in particular their applications to quantifying and uncovering dynamics in biological, chemical, and physical systems. 

\end{IEEEbiography}

\begin{IEEEbiography}[{\includegraphics[width=1in,height=1.25in,clip,keepaspectratio]{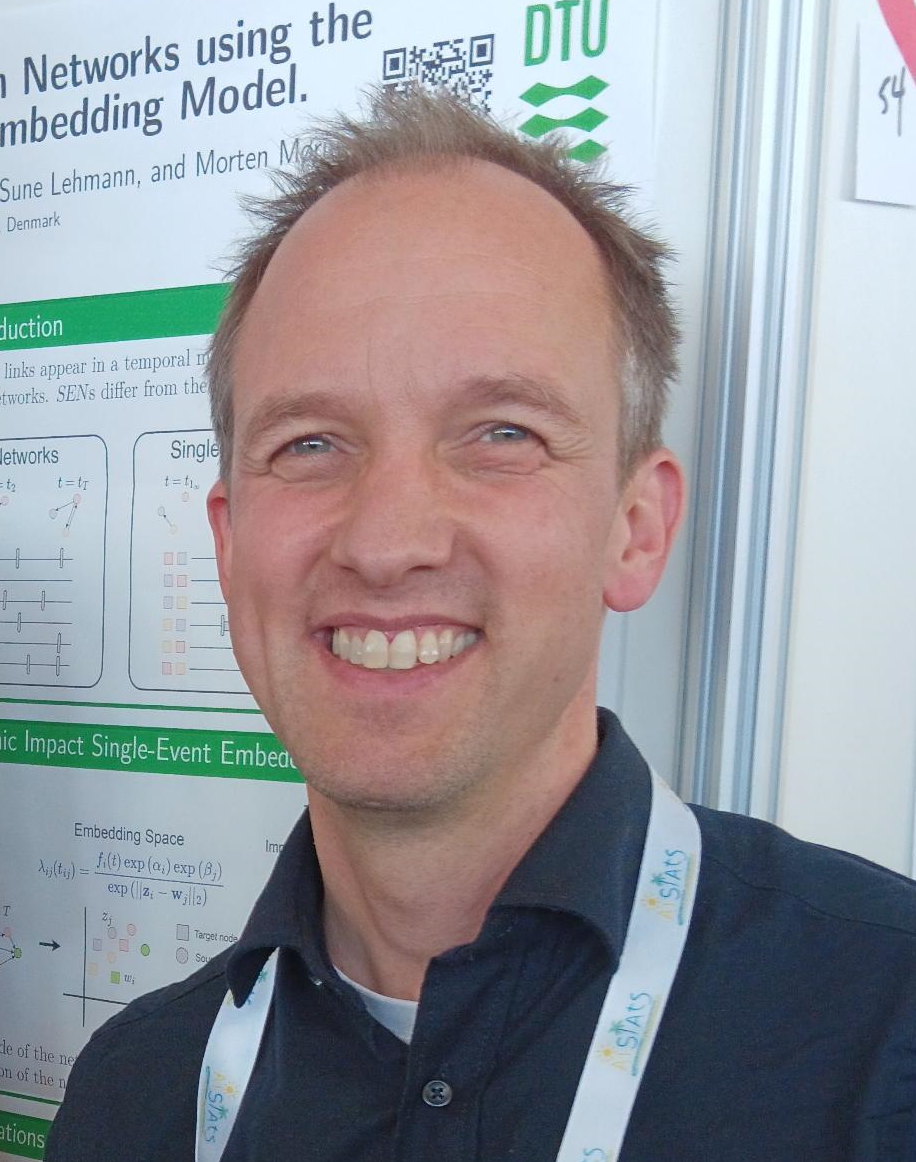}}]{Morten Mørup}
received his MS and PhD degrees in
applied mathematics from the Technical University of
Denmark, Denmark, where he is currently professor
of machine learning for the life-sciences at the
Section for Cognitive Systems at DTU Compute. He
has been associate editor of the IEEE Transactions
on Signal Processing, area chair for multiple leading machine learning conferences, and his research
focuses on machine learning, tensor decomposition,
and complex network modeling for the modeling of
life-science data.
\end{IEEEbiography}

\end{document}